\documentclass[a4paper]{article}
\usepackage[english]{babel}
\usepackage[utf8]{inputenc}
\usepackage[T1]{fontenc}
\usepackage[a4paper,top=3cm,bottom=2cm,left=3cm,right=3cm,marginparwidth=1.75cm]{geometry}
\usepackage{amsmath}
\usepackage{graphicx}
\usepackage[colorinlistoftodos]{todonotes}
\usepackage[colorlinks=true, allcolors=blue]{hyperref}
\usepackage{biblatex}
\usepackage{subfig}
\usepackage{float}
\usepackage[title]{appendix}
\usepackage{authblk}
\usepackage{blindtext}
\title{ViewPCL: a point cloud based active learning method for multi-view segmentation}
 
\author{Christian Hilaire\footnote{\texttt{Christian.Hilaire@proton.me}}, Sima Didari\footnote{\texttt{Sima.Didari@gmail.com}}}
\date{}
\newcommand{\argmax}{\arg\!\max}
\begin{document}
\maketitle

\begin{abstract}
We propose a novel active learning framework for multi-view semantic segmentation. This framework relies on a new score that measures the discrepancy between point cloud distributions generated from the extra geometrical information derived from the model's prediction across different views. Our approach results in a data efficient and explainable active learning method. The source code is available at \href{https://github.com/chilai235/viewpclAL}{https://github.com/chilai235/viewpclAL}.

\end{abstract}

\section{Introduction}

Acquiring high-quality ground truth data poses significant challenges for applying deep learning techniques in real-world scenarios. Active learning offers a promising solution to reduce labeling costs by actively selecting samples based on a policy and querying their labels. As a result, machine learning models can be trained using only a fraction of the data while achieving comparable performance.

One popular active learning strategy is uncertainty sampling. It focuses on requesting labels for samples that the model is most uncertain about, maximizing overall model improvement. For multi-view scenarios, multiple images of the same scene in different view angles exist. Thus, inherently  extra information can be extracted from the set of images to further enhance the uncertainty measure used in the active learning frameworks. The ViewAL \cite{viewal} active learning method is the first uncertainty sampling method to propose an uncertainty score that leverages the geometric constraints across different views: a surface point in a scene should receive consistent labeling across different views. ViewAL also implemented an active learning policy that selects superpixels to be labeled instead of whole images. Hence, uncertainty scores are superpixel uncertainty scores that are computed by aggregating pixel-wise defined uncertainty scores.

In our work, we introduce an uncertainty sampling based active learning method. Like ViewAL, it considers the geometric constraints across different views and also uses a superpixel selection policy. But unlike ViewAL, our method leverages a new superpixel score that is not a simple aggregation of pixel-wise defined uncertainty scores. This uncertainty score takes into account the metric space structure of the superpixel and, in that sense, our approach is more geometric-driven. In summary, our contributions are:

\begin{itemize}

\item Presenting a new uncertainty score for multi-view active learning semantic segmentation.  This score measures the discrepancy between point cloud distributions generated from the extra geometrical information derived from the model's prediction across different views.

\item A superpixel-based selection policy that reduces the labeling effort while preserving annotation quality.

\end{itemize}

\section{Related work}

Active learning has emerged as a fundamental approach to address annotation challenges in deep learning by strategically selecting the most informative samples for labeling \cite{settles1}. The field encompasses three primary groups of approaches: 
\begin{itemize}
\item \emph{uncertainty-based approaches} that query samples that the learning algorithm is most uncertain of. These approaches include methods based on entropy measures \cite{settles2,hwa2004, wang2016cost, joshi2009multi, gal2017deep}, margin-based methods, loss prediction \cite{yoo2019learning} methods, and Bayesian active learning methods with a mutual information acquisition function \cite{bald, batchbald}. There are also committee-based methods \cite{seung1992query, mccallum1998employing, dagan1995committee} and ensemble methods \cite{beluch2018power,chitta2018large}.
\item \emph{diversity-based methods} that target samples that are the most representative of the input distribution. Methods that fall into this category are clustering methods \cite{nguyen2004active,xu2007incorporating}, core-set selection methods \cite{sener2017active} and gradient embedding techniques \cite{badge}.
\item \emph{expected model change techniques} that prioritize samples that lead to significant model updates \cite{craven2008multiple,freytag2014selecting,kading2016active,vezhnevets2012active}. 
\end{itemize}
The groups listed above are not mutually exclusive. For example, hybrid methods that combine uncertainty and diversity signals have also been developed \cite{yang2017suggestive}.

Semantic segmentation utilizing deep neural networks \cite{long2015fully, ronneberger2015u, badrinarayanan2017segnet, Deeplab, zhao2017pyramid} poses significant computational and methodological challenges for active learning frameworks due to the high-dimensional pixel-wise labeling requirements inherent in image data. In the domain of multi-view semantic segmentation, these challenges are further amplified by the increased data volume from capturing multiple perspectives of each scene, necessitating extensive labeling across all viewpoints. To address these constraints, an uncertainty estimation approach leveraging view consistency principles \cite{viewal} has been introduced assuming that surface point in a scene should be labeled consistently across different views. Active learning methods based on view consistency have also been developed for video recommendation \cite{cai2019multi}, web page classification \cite{muslea2006active}, and image classification \cite{zhang2009multi}. These methods demonstrate the strength of consistency-based approaches in active learning strategies.

Our ViewPCL method is part of the family of view consistency based methods. It builds on the approach presented in ViewAL \cite{viewal}. However, our method leverages more the metric space structure of superpixels, thus making it more geometric-driven.

\section{ViewPCL}
Similar to ViewAL, an active learning round of ViewPCL consists of four main steps:
\begin{enumerate}
\item The network is trained on the current set of labeled data.
\item The model uncertainty is estimated on the unlabeled part of the data. 
\item The estimated model uncertainty is used to select the next set of data to be labeled.
\item The new annotations are added to the set of labeled data.
\end{enumerate}
The method selects regions of images to be labeled instead of whole images. These regions are generated in advance as \emph{superpixels}. The estimated model uncertainty will be computed in terms of a score for each superpixel. These \emph{superpixel scores} measure the inconsistency in the prediction based on different views.

\subsection{The cross-projected class probability maps}
 For each pixel $z$ of an image $I_i$, the trained segmentation network provides a class probability map $P_{i}^{z}$: for each class/label $c$, $P_{i}^{z}(c)$ is the probability that the pixel belongs to the class. Using pose and depth information, pixels from an image can be back-projected in 3D space and then projected onto other images. This allows us to cross-project class probability maps to other images.  Given images $I_i$ and $I_j$ such that the cross-projection of $I_{j}$ onto the plane containing $I_i$ actually intersects the image $I_i$ in a region $\mathcal{U} \subset I_{i}$, we will denote by $P_{ij} =\{P_{ij}^{z} \}_{z \in \mathcal{U}}$ the family of class probability maps cross-projected from $I_{j}$ onto $I_{i}$: $P_{ij}^{z} = P_{j}^{w}$ where $w \in I_{j}$ is the pixel that cross-projects onto $z \in \mathcal{U}$.

 \subsection{The induced point cloud distribution of a class}
Suppose we are given a region $\mathcal{D}$ of an image $I_{i}$ and a family of class probability maps $P = \{P^{z} \}_{z \in \mathcal{D}}$. In addition, we assume that we are given a \emph{selection} distribution $q$ on $\mathcal{D}$. Given $z \in \mathcal{D}$, $q(z)$ represents the probability that $z$ is a pixel of interest.

For a given class $c$ and a pixel $z \in \mathcal{D}$, the value $P^{z}(c)$ is a conditional probability of observing the class $c$ given that the pixel $z$ is selected.
Using the chain rule of probability, we can compute the associated probability $p^{q}(c)$ of observing the class $c$ in the region $\mathcal{D}$.
\begin{align*}
p^{q}(c) = \sum_{z  \in \mathcal{D} } P^{z}(c) \cdot q(z).
\end{align*}
Assuming $p^{q}(c) > 0$, we can also compute the conditional probability $p^{q}(z|c)$ of selecting a pixel $z$ given that the class $c$ is observed. This defines a distribution 
\begin{eqnarray*}
\mu^{q, c} = \sum_{z \in \mathcal{D}} w_{z} \delta_{z}
\end{eqnarray*}
on $\mathcal{D}$ where $w_{z} = p^{q}(z|c)$ and $\delta_{z}$ is the Dirac delta measure at $z$. This is the associated \emph{point cloud distribution of the object} $c$ in $\mathcal{D}$.

Suppose we have two families of class probability maps $P_1$ and $P_2$ on $\mathcal{D}$.  Clearly, if the two families $P_1$ and $P_2$ are similar, then the point cloud distributions $\mu_{1}^{q, c}$ and $\mu_{2}^{q, c}$ should also be similar for a given class $c$. A popular way of measuring the discrepancy between two point cloud distributions on a metric space is in terms of a Wasserstein distance or an approximation of a Wasserstein distance \cite{santambrogio2015optimal}.

The Wasserstein distance between two distributions $\mu$ and $\nu$ measures the minimal cost of reshaping the distribution $\mu$ into the distribution $\nu$. More precisely, given two point cloud distributions $\displaystyle \mu = \sum_{i =1 }^{m} a_i \delta_{x_i}$ and $\displaystyle \nu =  \sum_{j =1}^{n} b_j \delta_{y_j}$ on a finite metric space $(X,d)$, the Wasserstein distance of order $p$, $W_{p}(\mu, \nu)$, is the solution of the optimization problem:

\begin{eqnarray*}
W_{p}^{p}(\mu, \nu) = \inf_{\pi \in \Pi(\mu, \nu) } \sum_{i = 1}^{m} \sum_{j = 1}^{n} \pi_{ij} d(x_i, y_j)
\end{eqnarray*}
where $\pi = \{\pi_{ij} \}_{ 1 \leq i \leq m, 1 \leq j \leq n}$ is a \emph{transport plan}: for each state $x_i$ of $\pi$ and each  state $y_j$ of $\nu$, $\pi_{ij}$ is the probability mass transported from $x_i$ to $y_j$. The optimal transport plan must then satisfy the constraints $\pi_{ij} \geq 0$, $\displaystyle \sum_{j = 1}^{n} \pi_{ij} = a_{i}$ and $ \displaystyle \sum_{i = 1}^{m} \pi_{ij} = b_{j}$ for each $i$ and $j$. It is well known that this optimization problem has a solution and that it defines a distance function on the space of probability distributions of the metric space.

Measuring the dissimilarity of the families of class probability maps $P_{1}$ and $P_{2}$ in terms of the Wasserstein distance between generated point cloud distributions $\mu_{1}^{q, c}$ and $\mu_{2}^{q, c}$ presents the following advantages: 
\begin{itemize}
    \item It is a dissimilarity score that takes into account the geometry of the subregion $D$.
    \item It can be given a simple interpretation: it measures how badly $P_{1}$ and $P_{2}$ disagree on the configuration of the object $c$ in the subregion $\mathcal{D}$.
\end{itemize}

The ViewPCL active learning  method  uses a superpixel score, the \emph{viewpcl inconsistency score}, that is defined in terms of such dissimilarity measures.

\subsection{Viewpcl inconsistency score}
Given a superpixel $\mathcal{R}$ of an image $I_{i}$, the computation of its viewpcl inconsistency score involves the following steps.

\subsubsection{Selection of ``overlap subregions'' of $\mathcal{R}$}
 We consider images $I_{j}$ of the dataset such that the cross-projection onto $I_{i}$ intersects the region $\mathcal{R}$. For computational efficiency, we set a lower limit on the allowed relative size of the overlap. Given one of these images $I_{j}$, we take $\mathcal{D} \subset \mathcal{R}$ to be the corresponding \emph{overlap region}, i.e the intersection of $\mathcal{R}$ with the cross-projection.  We will compute for this subregion $\mathcal{D}$ a dissimilarity score $s(\mathcal{D})$ based on two families of class probability maps: the first one being the family of class probability map $P_i$ of the underlying image $I_{i}$ and the second being the cross-projected family $P_{ij}$ generated from the cross-projection of the class probability maps of $I_{j}$ onto $I_{i}$. For convenience, we will denote these families of class probability maps by $P_1$ and $P_2$ respectively.

\subsubsection{Computation of a score on each subregion $\mathcal{D}$}
Suppose we are given one of these subregion $\mathcal{D}$ and its associated pair $(P_1, P_2)$ of families of class probability maps. The score $s(D)$ will measure the dissimilarity between induced point cloud distributions. 
There are several reasonable choices for the selection distribution $q$, but we will simply set it to be the uniform distribution on $\mathcal{D}$. For the choice of the classes $c$, we propose to look at the most ``prominent ones''. More precisely, we first determine the most probable class for the family $P_{1}$
\begin{align*}
\hat{c}^{1} = \argmax_{c} p^{q}_1(c)
\end{align*}
and we consider the associated point cloud distributions $\mu_{1}^{q, \hat{c}^{1}}$ and $\mu_{2}^{q, \hat{c}^{1}}$. At a minimum, we expect the families $P_1$ and $P_2$ to agree on the configuration of the object $\hat{c}^{1}$. Our score will then involve a dissimilarity measure between the point cloud distributions $\mu_{1}^{q, \hat{c}^{1}}$ and $\mu_{2}^{q, \hat{c}^{1}}$. In addition, we take into account that the most probable label $\hat{c}^{2}$ for the family $P_{2}$ might not be equal to $\hat{c}^{1}$. The score $s(\mathcal{D})$ will be given by:
\begin{eqnarray*}
s(\mathcal{D}) = \frac{W(\mu_{1}^{q, \hat{c}^{1}}, \mu_{2}^{q, \hat{c}^{1}})+ W(\mu_{1}^{q, \hat{c}^{2}}, \mu_{2}^{q, \hat{c}^{2}})}{2}
\end{eqnarray*}
where $W$ denotes the Wasserstein distance of a given order or some approximation. It is an average of the discrepancy in the configuration of object $\hat{c}^{1}$ and the discrepancy in the configuration of object $\hat{c}^{2}$.  As mentioned in the ViewAL paper, a superpixel will often contain a small number of object-classes. Considering only the most probable classes should be reasonable for the first few rounds of active learning. 

Finally, our definition here assumes that each point cloud distribution exists. For instance, to compute $W(\mu_{1}^{q, \hat{c}^{1}}, \mu_{2}^{q, \hat{c}^{1}})$, the condition $p_{2}^q(\hat{c}^{1}) > 0$ must hold. In the rare cases that this condition does not hold, we compute a measure defined in terms of a transport plan that ``moves'' the point cloud $\mu_{1}^{q, \hat{c}^{1}}$ outside of the subregion $\mathcal{D}$. Details are provided in the Appendix. We will abuse notation and denote this measure by $W(\mu_{1}^{q, \hat{c}^{1}}, \mu_{2}^{q, \hat{c}^{1}})$ as well.

\subsubsection{Computation of the superpixel score $s(\mathcal{R})$ }
The score for the superpixel $\mathcal{R}$ is given by a simple aggregation of the scores of its overlap regions:
\begin{align*}
s(\mathcal{R}) = \frac{1}{Z}\sum_{\mathcal{D}}  w(\mathcal{D})\cdot s(D)
\end{align*}
where $w(\mathcal{D}) = \frac{|\mathcal{D}|}{|\mathcal{R}|}$ is the relative size of the region and $Z = \sum w(\mathcal{D})$ is the normalization term. This score is the viewpcl inconsistency score of the superpixel $\mathcal{R}$.

\subsection{Region selection policy}
The ViewPCL region selection policy will be based on two scores: 
\begin{itemize}
\item The view divergence superpixel score of viewAL.
\item Our proposed viewpcl inconsistency score.
\end{itemize}
Our selection policy will be a simple modification of the viewAL selection policy. Given a pool
\begin{eqnarray*}
C = \{ (j,\mathcal{S}) | \mathcal{S} \text{ superpixel of image } I_{j} \}
\end{eqnarray*}
of candidate superpixels to be annotated, we select the next superpixel  to be annotated by executing the steps below.
\begin{enumerate}
\item We first select the superpixel $(i,\mathcal{R})$ with the highest viewpcl inconsistency score:
\item We then identify the subset $C_{\mathcal{R}}$ of $C$ consisting of  candidate superpixels whose cross-projection onto the image $I_{i}$ intersects the superpixel $\mathcal{R}$ in a subregion $\mathcal{D} \subset \mathcal{R}$ with relative size greater than a user specified threshold. Just like in viewAL, the superpixel selected for annotation is the superpixel of $C_{\mathcal{R}}$ with the highest view divergence score and the remaining superpixels in $C_{\mathcal{R}}$ are removed from further selection considerations.
\end{enumerate}
We will show that this simple modification of the viewAL selection algorithm already gives measurable improvements.

\section{Experiments setup}
We evaluate our ViewPCL method by running experiments where active learning rounds are executed until the labeling budget is reached or all the data is labeled. \footnote{Source code available at \href{https://github.com/chilai235/viewpclAL}{https://github.com/chilai235/viewpclAL}.}
 
\subsection{Network architecture}
We use DeepLabv3+ \cite{Deeplab} with MobileNetv2 \cite{MobileNetV2} network architecture. This combination allows fast training and low memory requirement for inference. MobileNetv2 backbone weights are initialized by weight of the pre-trained model on ILSVRC 1000-class classification \cite{ilsvrc} and Kaiming initialization is used for the remaining layers \cite{kaiming}.  Blur, random crop, random flip and Gaussian noise data augmentation are implemented at the training as well.

\subsection{Dataset}
We evaluate our approach on the public dataset SceneNet-RGBD \cite{scenenet1, scenenet2}. It is a synthetic dataset composed of large-scale photorealistic renderings of indoor scene trajectories. We use the same subsets of images selected in \cite{viewal} for training and validation. These subsets are generated as follows:
\begin{itemize}
\item The SceneNet-RGBD training dataset is divided into 17 sets. The training set for our experiment was obtained by selecting every tenth frame from 5 of these sets ( sets 0, 4, 7, 14 and 16). This resulted in 72990 images for training.
\item Similarly, we obtained a validation set of 15000 images by selecting every tenth frame of the SceneNet-RGBD validation dataset.
\end{itemize}
We follow the same pre-processing steps outlined in \cite{viewal}. This includes generating superpixels for each image of our dataset. The SEEDs algorithm \cite{seeds} is used to generate 40 superpixels per image. In addition, we compute and save for each superpixel a collection of overlap regions generated by cross-projections. These overlap regions are used to compute the viewpcl inconsistency scores. For computational efficiency, we only consider overlap regions whose size relative to the superpixel is greater than or equal to a user specified threshold. 

\subsection{Training step}
At the start of each experiment, the training seed set is initialized with fully annotated $1.5 \%$ of training images that are randomly selected from the unlabeled dataset. For each active learning round, the networks are trained with SGD for 50 epochs. We use an initial learning rate of 0.0004. This value is decayed on the 35th epoch to 0.1 times the original value. The momentum is set to 0.9 and the weight decay penalty to 0.0005. SceneNet-RGBD does not have a test dataset. So we compute the mIoU on the validation set.

\subsection{Uncertainty scoring step}
To improve robustness of the probability estimates, we use the MC dropout method as suggested in \cite{viewal}. For an image $I_i$ of our unlabelled dataset and pixel $z \in I_{i}$, the class probability map that we use for our uncertainty scoring is given by:
\begin{eqnarray}
P_{i}^{z}(c) = \frac{1}{D} \sum_{d = 1}^{D} P_{i, d}^{z}(c)
\end{eqnarray}
where $D$ is the number of dropout runs of the network and $P_{i, d}^{z}(c)$ is the softmax probability of pixel $z$ belonging to class $c$ for the MC dropout run $d$.

After computing a point cloud distribution $\mu = \sum_{z} w_{z} \delta_{z}$ from a given class $c$, a given selection distribution $q$ and a given cross-projected class probability map, we perform some light processing to improve computational efficiency: we ignore points $z$ having weights lower than a certain threshold and we renormalize the weights for the remaining points. Finally, we use the Slice Wasserstein Distance implementation of the Python Library POT \cite{flamary2021pot} to compute the dissimilarity between point cloud distributions.

\subsection{Region selection step}
At the end of an active learning round, our proposed active learning algorithm is used to select superpixels equivalent to a requested number $K$ of images. For our experiments, we set $K = 1500$.

\section{Results} We compared our proposed active learning method to ViewAL. (For information about other active learning methods relevant to multi-view semantic segmentation, consult the appendix of \cite{viewal}.) For each method, we performed a total of 5 experiments. A different random seed was used to generate the seed set for each experiment. For 3 of the 5 experiments, a total of 7 active learning rounds were executed. For the remaining 2 experiments, we ran a total of 4 active learning rounds. 

\begin{figure}[htbp]
    \centering
    \subfloat[mIoU results]{\includegraphics[width=0.45\textwidth]{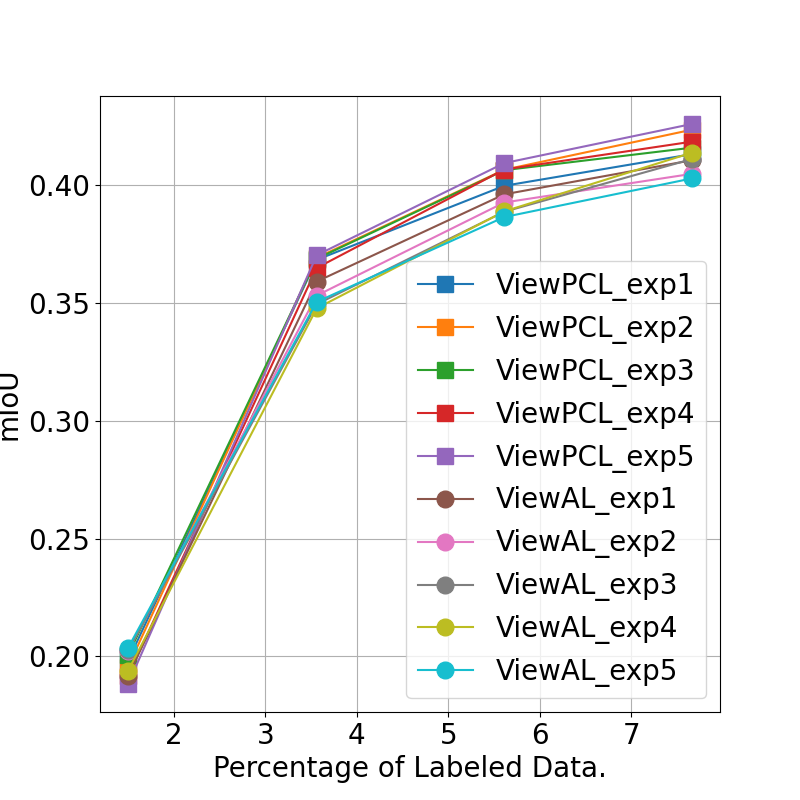}\label{fig:miou_a}}%
    \qquad
    \subfloat[Average mIoU results]{\includegraphics[width=0.45\textwidth]{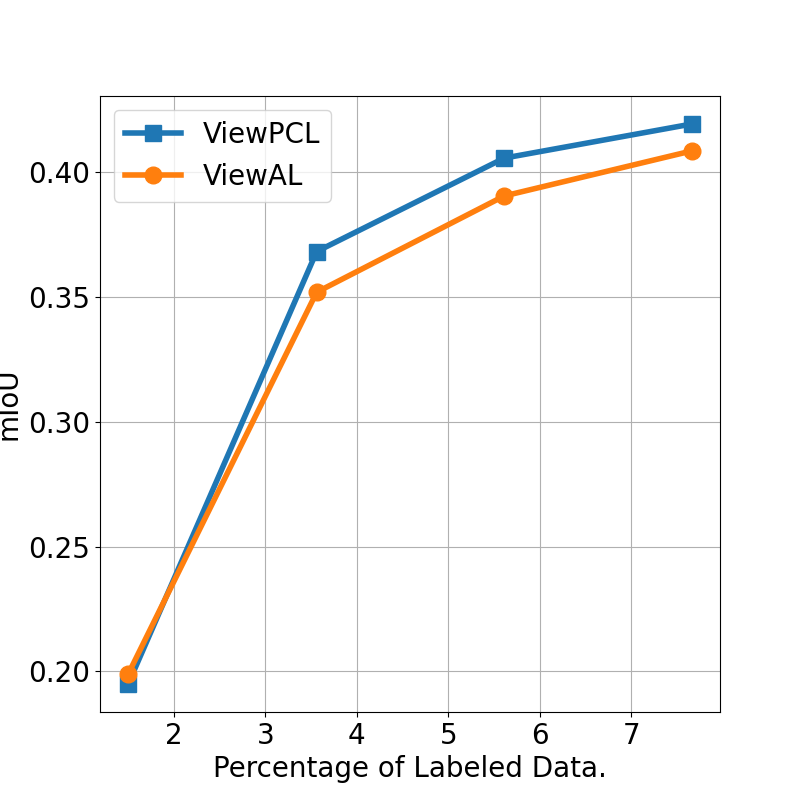}\label{fig:miou_b}}%
    \caption{mIoU results for 4 active learning rounds.}
    \label{fig:miou}
\end{figure}

The results of our experiments are shown in Figure~\ref{fig:miou}. In \ref{fig:miou_a}, the mIoU values over 4 active learning rounds for all the experiments are shown. In \ref{fig:miou_b}, we averaged the mIoU values over the 5 experimental values available for each method and each active learning round. Our proposed method clearly outperforms ViewAL on the SceneNet-RGBD dataset.

In the appendix, we provide equivalent plots for the subset of experiments that were executed for 7 active learning rounds.

\section{Conclusions}
We have introduced ViewPCL, a novel active learning method for semantic segmentation. The method selects superpixels to be labeled instead of whole images. We introduced the \emph{viewpcl inconsistency score}, a geometric-driven and interpretable metric for measuring inconsistencies between different views of a scene. We showed that our method provides measurable improvements in data efficiency when tested on the SceneNet-RGBD dataset. The definition of our viewpcl inconsistency score comes with some flexibility: the point cloud distributions generated for a label $c$ depend on our choice of the selection distribution $q$  for the overlap subregion $\mathcal{D}$ of a superpixel $\mathcal{R}$. 

Future research directions include exploring selection distributions $q$ that differ from the uniform distribution employed in this work. For instance, a Bayesian-like approach could employ a selection distribution based on the point cloud distributions from the previous active learning iteration. Additionally, our selection policy represents a straightforward modification of ViewAL's approach: we replaced ViewAL's view entropy score with our viewpcl inconsistency score. Developing more sophisticated selection policies that better leverage the viewpcl inconsistency score presents another promising avenue of research.

%\bibliographystyle{unsrt}
%\bibliography{main}
\printbibliography %Prints bibliography

\begin{appendices}

\section{Computing $W(\mu_{1}^{q, c}, \mu_{2}^{q, c})$  when $p_{2}^q(c) =  0$ }
Suppose we have a subregion $\mathcal{D}$ of a superpixel, a selection distribution $q$, and two families of class probability maps $P_1$ and $P_2$. Suppose as well we have a class $c$ such that $p_{1}^q(c) > 0$ but $p_{2}^q(c) =  0$. So there is a point cloud distribution $\mu_{1}^{q, c}$, but the point cloud distribution $\mu_{2}^{q, c}$ does not exist in $\mathcal{D}$. In this case, we compute a dissimilarity measure that is still defined in terms of a transport plan. Based on the family $P_2$, the ``object'' $c$ does not lie in the subregion $\mathcal{D}$. So we define a transport plan that will move the point cloud distribution $\mu_{1}^{q, c}$ outside of $\mathcal{D}$. More precisely, we consider the smallest bounding box that encloses the subregion ${D}$ and we compute the cost of the transport plan that moves each state $z$ of the point cloud distribution $\mu_{1}^{q, c}$ to the closest point on the boundary of the bounding box. This value will still be denoted by $W(\mu_{1}^{q, c}, \mu_{2}^{q, c})$.

\section{ViewAL active learning method}
We provide here a quick review of the ViewAL active learning method. 

For each pixel $z$ of an image $I_{i}$, we consider the set $\Omega_{i}^{z}$ of all class probability maps at $z$ that are obtained from cross-projections.
\begin{align*}
\Omega_{i}^{z} = \{ P_{j}^{w} | w \text{ of image } I_{j} \text{ cross-projects to } z \}
\end{align*}
Two scores are then derived from this set $\Omega_{i}^{z}$: the \emph{view entropy score} and the \emph{view divergence score}.

\subsection{View entropy score}
Given a pixel $z$ of image $I_{i}$ and its associated set of class probability maps $\Omega_{i}^{z}$, we compute the cross-projected class probability map $Q_{i}^{z}$
\begin{align*}
Q_{i}^{z}(c) = \frac{1}{|\Omega_{i}^{z}|}\sum_{P \in \Omega_{i}^{z}} P(c)
\end{align*}
The view entropy score is then given by the entropy of this distribution:
\begin{align*}
VE_{i}^{z} = -\sum_{c} Q_{i}^{z}(c) \log \left(Q_{i}^{z}(c)\right).
\end{align*}

\subsection{View Divergence score}
The view divergence score for a pixel $z$ of image $I_{i}$ measures an average discrepancy between its class probability map $P_{i}^{z}$ and the class probability maps of the family $\Omega_{i}^{z}$.
\begin{align*}
VD_{i}^{z} = \frac{1}{|\Omega_{i}^{z}|}\sum_{P \in \Omega_{i}^{z}} D_{KL}(P_{i}^{z}|| P)
\end{align*}
where $D_{KL}$ denotes the KL-divergence measure.

Superpixel scores are computed from the pixel-wise view entropy and view divergence scores with a simple averaging operation: for a superpixel $\mathcal{R}$ of an image $I_{i}$ the superpixel view entropy and view divergence scores are given by:
\begin{align*}
VE_{i}^{\mathcal{R}} = \frac{1}{|\mathcal{R}|} \sum_{z \in \mathcal{R}} VE_{i}^{z}
\end{align*}
and  
\begin{align*}
VD_{i}^{\mathcal{R}} = \frac{1}{|\mathcal{R}|} \sum_{z \in \mathcal{R}} VD_{i}^{z}
\end{align*}

\subsection{Region selection policy}
ViewAL is a supepixel selection method. Given a set $C$ of candidate superpixels, the next superpixel to be annotated is obtained by first selecting from $C$ the superpixel $\mathcal{R}$ with the highest view entropy. 
We then determine the subset $C_{\mathcal{R}}$ of $C$ consisting of candidate superpixels whose cross-projection onto the image $I_{i}$ intersects the superpixel $\mathcal{R}$ in a subregion $\mathcal{D} \subset \mathcal{R}$ with relative size greater than a specified threshold. The superpixel selected for annotation is the superpixel of $C_{\mathcal{R}}$ with the highest view divergence score. The remaining superpixels in $C_{\mathcal{R}}$ are removed from the set of candidates.

\section{Additional plots}

\begin{figure}[htbp]
    \centering
    \subfloat[mIoU results]{\includegraphics[width=0.45\textwidth]{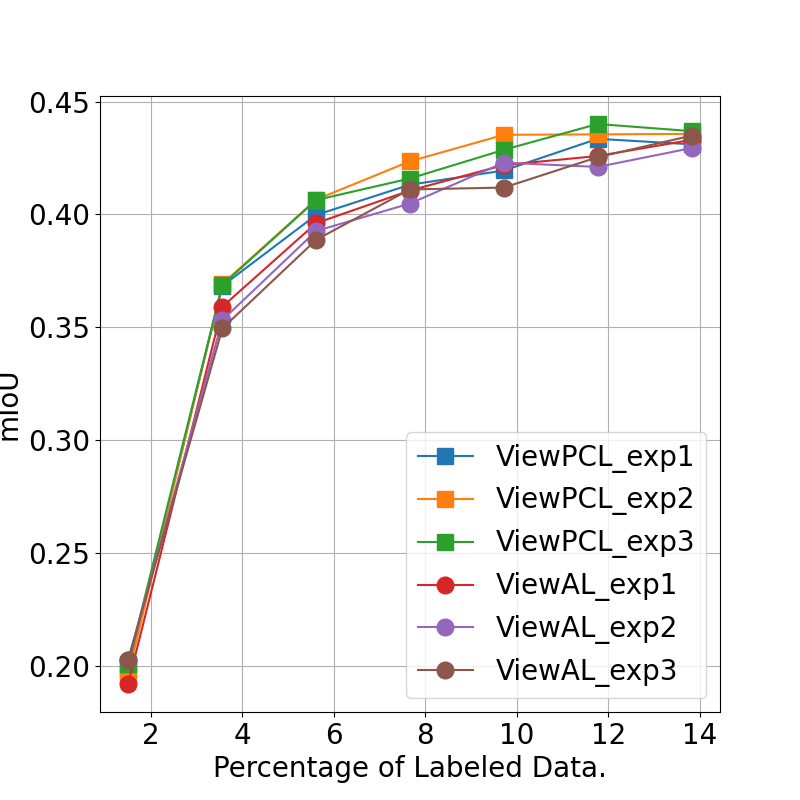}}\label{fig:miou7_a}%
    \qquad
    \subfloat[Average mIoU results]{\includegraphics[width=0.45\textwidth]{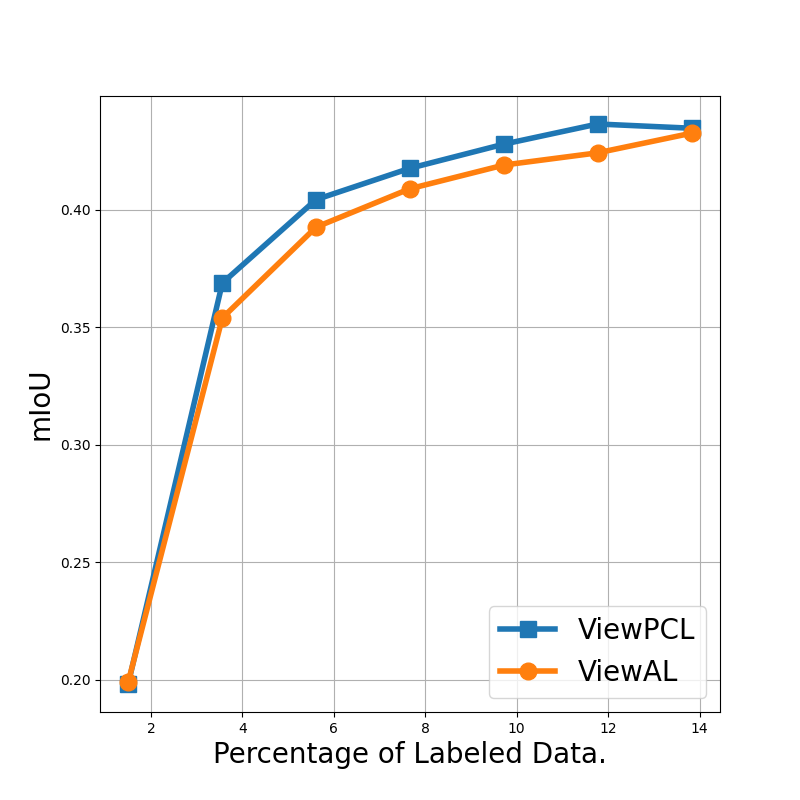}}\label{fig:miou7_b}%
    \caption{mIoU results for 7 active learning rounds.}
    \label{fig:miou7}
\end{figure}

\begin{figure}[H]
    \centering
    \subfloat[Training loss]{\includegraphics[width=0.45\textwidth]{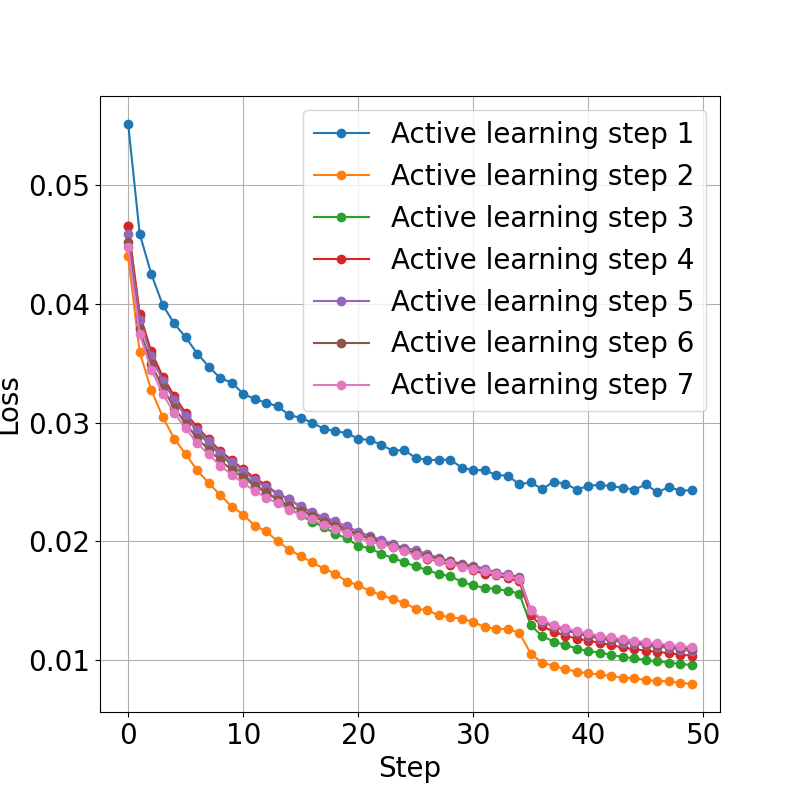}}\label{fig:train_loss}%
    \qquad
    \subfloat[Validation loss]{\includegraphics[width=0.45\textwidth]{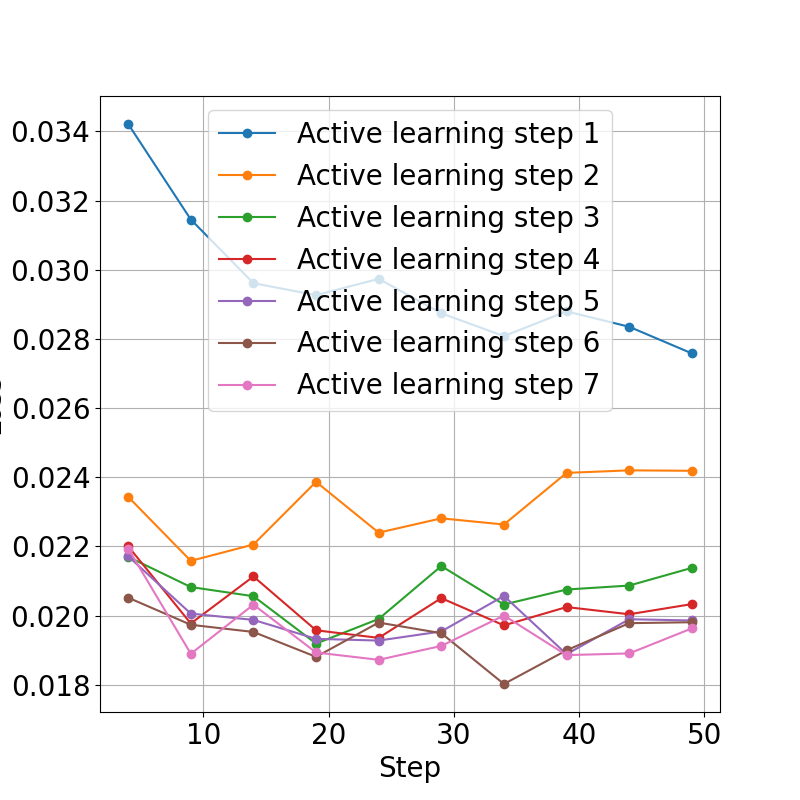}}\label{fig:miou7_b}%
    \caption{Loss values for one of the ViewPCL experiment showing convergence of the model.}
    \label{fig:miou7}
\end{figure}

\end{appendices}

\end{document}